\documentclass{article} % For LaTeX2e
\usepackage{nips12submit_e,times}
\usepackage{amsmath}
\usepackage{amssymb}
\usepackage{nicefrac}
\usepackage{graphicx,subfigure}

\title{Gradient Driven Learning for Pooling in Visual Pipeline Feature Extraction Models}

\author{
Derek Rose and Itamar Arel\\
Deptartment of Electrical Engineering and Computer Science\\
University of Tennessee\\
\texttt{derek@utk.edu, itamar@ieee.org} \\
}

\nipsfinalcopy % Uncomment for camera-ready version

\begin{document}

\maketitle

\begin{abstract}
Hyper-parameter selection remains a daunting task when building a pattern recognition architecture which performs well, particularly in recently constructed visual pipeline models for feature extraction. We re-formulate pooling in an existing pipeline as a function of adjustable pooling map weight parameters and propose the use of supervised error signals from gradient descent to tune the established maps within the model. This technique allows us to learn what would otherwise be a design choice within the model and specialize the maps to aggregate areas of invariance for the task presented. Preliminary results show moderate potential gains in classification accuracy and highlight areas of importance within the intermediate feature representation space.
\end{abstract}

\section{Introduction}
Multi-stage visual pipelines for learning feature representations of images have recently proven valuable for classifying small objects from a variety of lighting conditions, scales, and poses. An effective variant of these was explored
in experiments by \cite{coates_analysis_2011} that compared features learned and encoded by stages that originate in encoder-decoder networks, deep learning \cite{bengio_learning_2009}, and bags of features models \cite{lazebnik_beyond_2006}. This architecture partitions images into patches to perform local learning of an over-complete codebook and uses this codebook to form global representations of the images for classification. Intermediate or mid-level representations are of high-dimensionality and retain the spatial structure of the image. Pooling is an integral late stage that performs the same role as the sub-sampling layer of a convolutional neural network (CNN): reduction in the final number of features (passed to the classifier or next layer) by an aggregation operation meant to improve invariance to small changes \cite{lecun_gradient-based_1998}. 

Unfortunately, each additional stage of the architecture often adds hyper-parameters for model selection that must be explored. For the pooling layer the number of pools, their structure or spatial layout, weights within this region, and operator (often max, average, or $p$-norm) are hyper-parameters that are frequently chosen by rules of thumb. Recently, \cite{jia_beyond_2012} explored pool selection by optimization over a full training set with an over-complete number of pools and achieved excellent improvements over standard pyramid models, although their method uses a feature selection stage with retraining to tractably create a classifier. In this paper we present a method for learning pooling maps with weight parameters that may optimize or tune the feature space representation for better discrimination through stochastic gradient descent. This converts two of the model choices above to parameters which may be learned from a limited stream of labeled training data.

Back-propogation through the architecture in \cite{coates_analysis_2011} is used to obtain the appropriate weight updates. This technique stems back at least to the inception of convolutional neural networks and graph transformer networks  \cite{lecun_gradient-based_1998}, where each module or layer of the network may be utilized in a forward pass output calculation and backward pass parameter update. 

\section{Architecture Description and Design}
% Problem statement
The multi-stage image recognition architecture considered from \cite{coates_analysis_2011} consists of patch extraction, normalization and whitening, codebook learning, feature encoding, spatial pooling, and classification. As guided by \cite{coates_analysis_2011} we use hyper-parameters such as the number of codewords, $k=400$, patch size $w=6$ pixels, stride$=1$ for dense extraction, triangle encoding, and use of patch whitening and pre-processing with a focus on the CIFAR-10 dataset. Quadrant based average pooling was used to sub-sample the intermediate feature space representation down to $4k$. This choice stemmed from spatial pyramids and was not a heavy focus for  \cite{coates_analysis_2011}, although motivating work in \cite{jia_beyond_2012} has shown there are both a large number of options for pooling regions and much performance to be gained. 

Although building pools for invariance can be partially based on intuition when patch extraction is performed spatially relevant to the original image (as a single layer in the architecture considered), pooling region choice for layers of features that build upon those generated in the lowest layer is not as straightforward (as noted by \cite{coates_learning_2012}). Special care must be taken to reorder features and assemble pools that maintain the structure inherent in the image. It would be beneficial if the region could change relative to problem demands as well as layer context, and we propose the use of pooling weight maps that may be adjusted as learning proceeds with gradient descent.

\section{Stochastic Gradient Descent Based Method for Weighted Pooling}
% Explanation of solution
To obtain continuous updates as well as a gradient signal we replace the support vector machine originally used for classification in \cite{coates_analysis_2011} with a feed-forward neural network with single hidden layer and mean-square error cost function. The outputs are one-hot binary vectors for the $t=10$ target classes and the inputs are pooled features $\mathbf{h}$. If we have a $n\times n$ pixel square image and let $P=\left(n(w-1)+1\right)$, with dense patch extraction we obtain a $P\times P\times k$ mid-level representation (this may be modified for non-square images). Pooling reduces this representation to $p\times k$ that can be flattened to form $\mathbf{h}$. We then extend the network back to include an additional input layer that holds our $p$ weight maps with size $P\times P$ that we denote as $\mathbf{W}^i$. The inputs to this new layer are encoded features $g$ computed from the $P\times P\times k$ patches.

% Similarity to CNN
This method shares many similarities with the sub-sampling layer in a CNN. In the sub-sampling layer each neuron receives the average of the features from the prior layer in its receptive field or pool (receptive fields of units do not overlap). Each unit has a coefficient and bias that are trained with gradient information and feed into a sigmoid activation function that controls the response of the sub-sampling unit \cite{lecun_gradient-based_1998}. Aside from its application in an alternate context, our approach differs in that the features which are averaged are no longer restrained to being equally weighted and multiple pools may utilize the same encoded patches. Learned weights combine encoded features $g$ from the mid-level representation and pass these through a linear activation with unity weight to a neural network classifier, i.e. pooled inputs $\mathbf{h}$ over encoded patches $\mathbf{g}$ in pool $i$ are computed\footnote{Codeword index has been omitted for clarity. $h$ and $g$ have a $k$-size dimension which $\mathbf{W}$ is shared over.} by
% Weighted Pooling:
\begin{equation}
h_{i} = f_{pool}\left(\mathbf{g}\right)=\sum_{m=1}^{P}\sum_{n=1}^{P}{\mathbf{W}_{m,n}^{i}*g_{m,n}}\label{eqn:weight_pool}.
\end{equation}
\begin{figure}[tb]
\begin{center}
%\subfigure[Mapped Connections]{%
\subfigure[]{%
	\label{fig:arch}
	\includegraphics[scale=0.52]{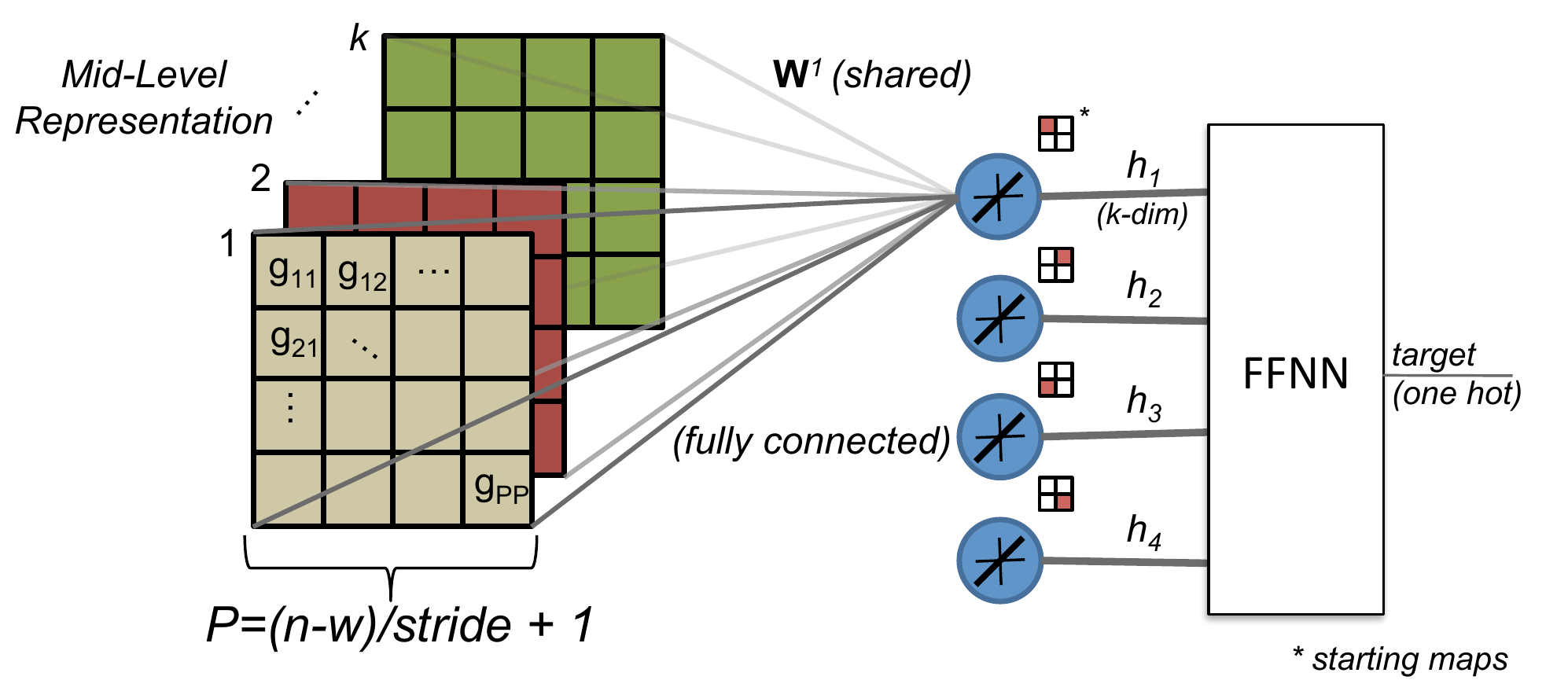}
}%	
%\subfigure[Learned Maps]{%
\subfigure[]{%
	\label{fig:maps}
	\includegraphics[scale=0.15]{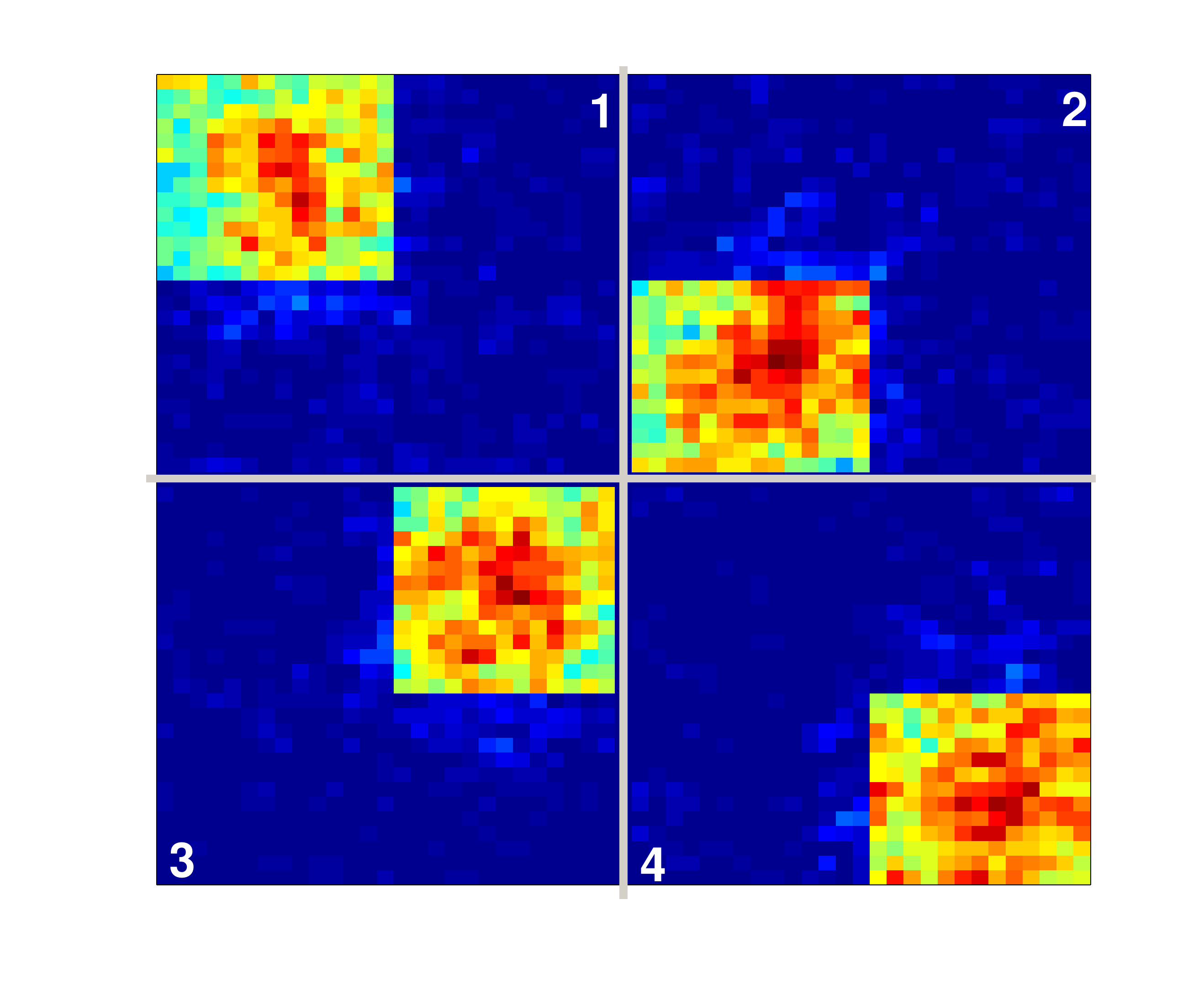}
}%
\caption{(a) Pool weight map connections and feed-forward net extension, (b) Learned weight maps}
\end{center}
\end{figure}
We currently have not explored using non-linear activation units with adjustable biases and coefficients and although we have no restriction that pools must be non-overlapping as in CNNs, we are currently examining modification to the loss function to maximize diversity among the pools. The weight sharing scheme employed by \cite{lecun_gradient-based_1998} for learning feature maps is similarly used here to reduce the parameter search space by sharing weights in codeword dimension $k$ of the mid-level representation. Figure \ref{fig:arch} illustrates the positioning of the learned parameters within the architecture. To replicate the quadrant based pooling we employ four maps that connect to every patch and are initially set to $W_{m,n}=1/{\left(P/2\right)^2}$ inside the hot zone of the corresponding quadrant map and zero elsewhere. After learning the codebook we train the original network with 128 hidden neurons and stochastic average mini-batch gradient descent (random batches of 10 images) while the pool weights are fixed. Network inputs are normalized to $\bar{h}$ post pooling with a mean and variance that is not modified. After a learning period we fix the classifier weights and learn the pooling weights. We have found this alternate training to perform better than simultaneous learning where the neural network can compensate for pooling weight changes.

\subsection{Weight Update Rule and Preliminary Results}
To update each pool $i$'s weights with loss function $J$ note we want $\Delta W_{m,n}^{i} = -\eta\nicefrac{\partial J}{\partial W_{m,n}^{i}} = -\eta\frac{\partial J}{\partial \bar{h}_i}\frac{\partial \bar{h}_i}{\partial W_{m,n}^{i}}$. From the original backpropagation pass we may create a new set of sensitivities $\boldsymbol{\delta}^{0} = -\nicefrac{\partial J}{\partial \bar{\mathbf{h}}} = \mathbf{v}^{1^{T}}\boldsymbol{\delta}^{1}$ ($v$ representing the original input to hidden weights) for the prepended network layer. Noting $\bar{h}_i = \nicefrac{\left(h_{i} - \mu_{i}\right)}{\sigma_i}$ and $\nicefrac{\partial\bar{h}_i}{\partial W_{m,n}^i}=\nicefrac{g_{m,n}}{\sigma_{i}}$ we obtain the update rule
\begin{equation}
\Delta W_{m,n}^{i} = \eta \delta_{i}^{0} \frac{g_{m,n}}{\sigma_{i}}\label{eqn:weight_update}.
\end{equation}
The pool weight gradient updates for the images in each mini-batch are averaged; in each, every codeword dimension $k$ contributes to the update as a sum. In preliminary tests we freeze network learning after presenting 250k examples from 80\% of the CIFAR-10 training set. The last validation accuracy before pool learning on the remaining 20\% averaged over five trials was 67.56\%. After training with an additional 15k training images and checking the validation set every 500 images, the average best post pool learning accuracy was 68.03\% for an improvement of $\mathbf{0.57}\%$. Figure \ref{fig:maps} contains an example of the weights learned that highlights the infinite number of possible pools gradient descent searches over.

%This model's stochastic update scheme allows us to continue to train and tune the parameters as inputs arrive which may be modeled as a non-stationary process where the distribution of variances within the images is varying.

This method may also be used to tune the choice of $p$ within the weighted $p$-norm.
Open issues remain, particularly in the learning rate choice (for which we have traded the pooling structure for hyper-parameter $\eta=5e^{-5}$) and the number of maps needed to cover separate areas of invariance. It would be preferred to select more than enough maps than necessary and later trim down redundant features, although we need to be careful to avoid overfitting the training set here (and in general for this approach). Unfortunately, each additional map adds $k$ inputs to the classifier. This problem relates closely to feature map count or hidden neuron count hyper-parameters in CNNs.

\bibliographystyle{IEEEtran}
{\footnotesize\bibliography{IEEEabrv,library}}

\end{document}